\documentclass[conference]{IEEEtran}

\IEEEoverridecommandlockouts

\usepackage{graphicx}
\usepackage{newlfont}
\usepackage{multicol}
\usepackage{multirow}
\usepackage[table]{xcolor}
\usepackage{tabularx}
\usepackage{cite}
\usepackage{amssymb}
\usepackage[cmex10]{amsmath}
\usepackage{array}
\usepackage{stfloats}

\usepackage{lipsum}

\usepackage{fancyhdr}
\graphicspath{{images/}{images/}}

\begin{document}

\title{Comparative Performance Analysis of Neural Networks Architectures on H2O Platform for Various Activation Functions}

\author{\IEEEauthorblockN{Yuriy Kochura$^{1, *}$, Sergii Stirenko$^2$, Yuri Gordienko$^2$, }
\IEEEauthorblockA{ $^1$Department of Physics of Energy Systems, $^2$Department of Computer Engineering\\ 
National Technical University of Ukraine "Igor Sikorsky Kyiv Polytechnic Institute"\\
Kyiv, Ukraine\\
$^*$iuriy.kochura@gmail.com}}

\maketitle

\thispagestyle{fancy}

\begin{abstract}
Deep learning (deep structured learning, hierarchical learning or deep machine learning) is a branch of machine learning based on a set of algorithms that attempt to model high-level abstractions in data by using multiple processing layers with complex structures or otherwise composed of multiple non-linear transformations. In this paper, we present the results of testing neural networks
architectures on H2O platform for various activation functions, stopping metrics, and other parameters of machine learning algorithm. It was demonstrated for the use case of MNIST database of handwritten digits in single-threaded mode that blind selection of these parameters can hugely increase (by 2-3 orders) the runtime without the significant increase of precision. This result can have crucial influence for opitmization of available and new machine learning methods, especially for image recognition problems.

\end{abstract}



\section{Introduction}

Machine learning (ML) is a subfield of Artificial Intelligence (AI) discipline. This branch of AI involves the computer applications and/or systems design that based on the simple concept: get data inputs, try some outputs, build a prediction. Nowadays, ML has  driven advances in many different fields~\cite{Kochura} like pedestrian detection, object recognition, visual-semantic embedding, language identification, acoustic modeling in speech recognition, video classification, fatigue estimation~\cite{Gordienko}, generation of alphabet of symbols for multimodal human-computer interfaces~\cite{Hamotskyi}, etc. This success is related to the invention of  more sophisticated machine learning models and the development of software  platforms  that  enable  the  easy  use  of  large  amounts of  computational  resources  for  training  such  models~\cite{Sibi}.

\begin{figure}[h!]
\centering
\includegraphics[width=3in]{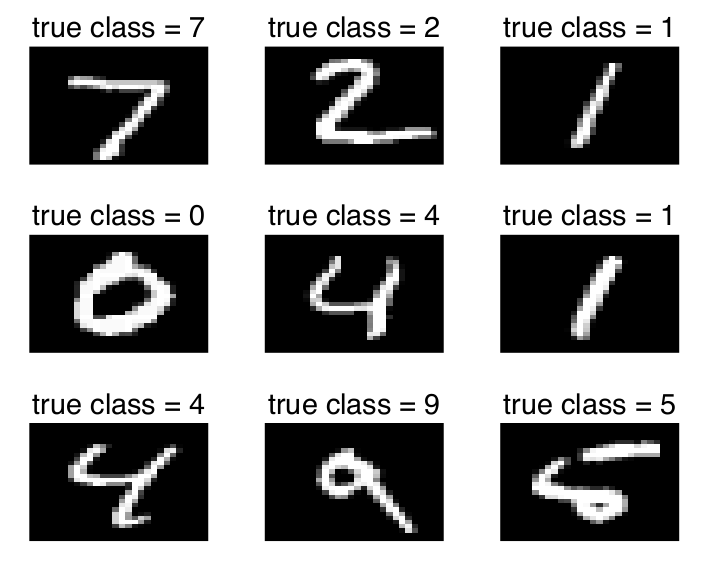}
\caption{Example Images of Handwriten Digits}
\label{fig1}
\end{figure}

H2O is one of such open source deep learning platform. We have tested the H2O system by using the publicly available MNIST dataset of handwritten digits. This dataset contains 60,000 training images and 10,000 test images of the digits 0 to 9. The images have grayscale values in the range 0:255.  Figure~\ref{fig1} gives an example images of handwritten digits that were used in testing. We have trained the net by using the host with Intel Core i7-2700K insight. The computing power of this CPU approximately is 29.92 GFLOPs.

In this paper, we present testing results of various net architectures by using H2O platform for single-threaded mode. Our experiments show that net architecture based on cross entropy loss function, tanh activation function, logloss and MSE  stopping metrics demonstrates better efficiency by recognition handwritten digits than other available architectures for the classification problem.

This paper is structured as follows. Section~\ref{Motivation} describes the motivation for this idea. Section~\ref{Common Information} introduces us with activation and loss functions and describes parameters of deep neural nets that were used in the experiments. In Section~\ref{Experimental Results}, we present our experimental results where we apply different activation functions and stopping metrics to the classification problem with use case in single-threaded mode. Section~\ref{Conclusion} contains the conclusions of the work.

\section{Motivation} \label{Motivation}
A motivation for providing testing of various neural network architectures on H2O platform has related to lacking publicly available similar publications, especially with an emphasize on influence of some parameters of machine learning method like activation function, stopping metrics, etc. This paper is a logical continuation of the previous paper~\cite{Kochura} and it aims to learn the main features of H2O platform for classification problem.

\section{Common Information} \label{Common Information}

\subsection{The Activation Functions}

\begin{figure}[h!]
\centering
\includegraphics[width=3in]{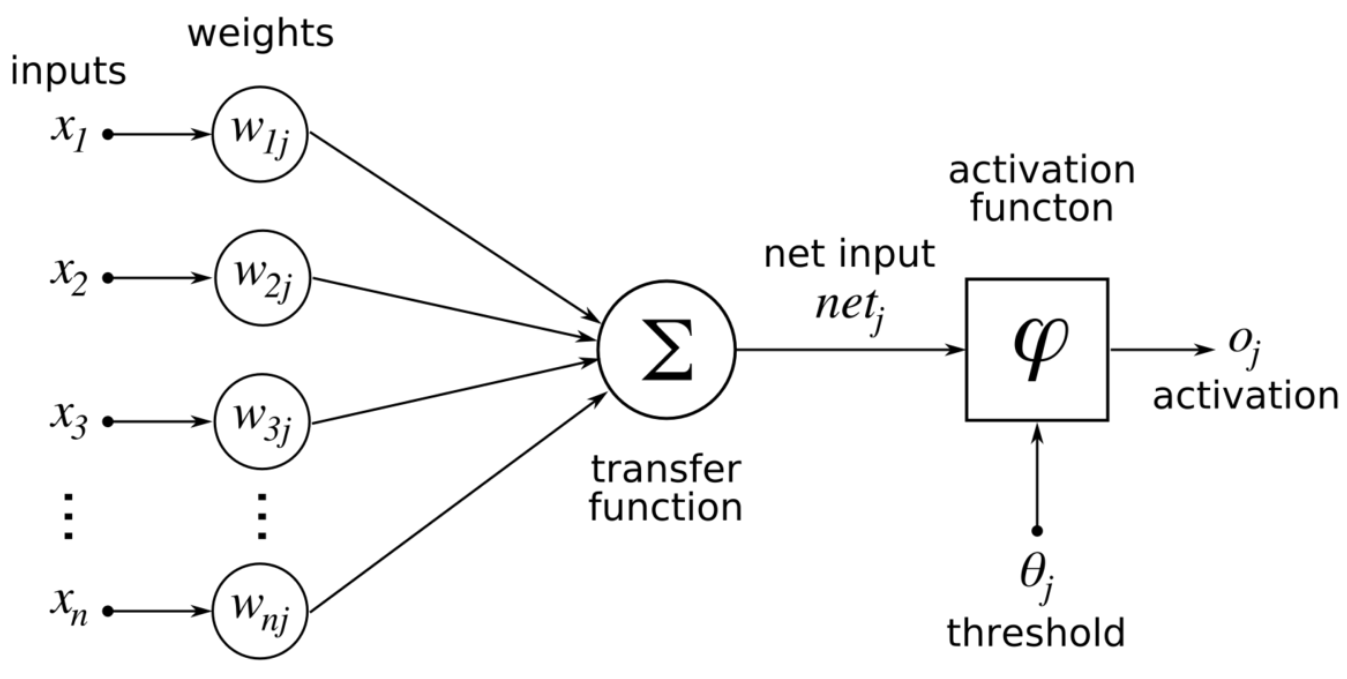}
\caption{The Role of Activation Function in the Process of Learning Neural Net~\cite{ANN}}
\label{fig1_1}
\end{figure}

Activation functions also known as transfer functions are used to map input nodes to output nodes in certain fashion~\cite{AF} (see the conceptual scheme of an activation function in Figure~\ref{fig1_1}). We are considering here most common activation functions that are widely using for deep learning.

The rectifier activation function is defined as 
\begin{equation}\label{rectifier}
f(x) = \max(0, x)
\end{equation}
where $x$ is the input to a neuron.

When given an input $x \in R^d$, a maxout activation function defins as folows:
\begin{equation}\label{maxout}
f_i(x) = \max_{j \in [1,k]}(x_{ij})
\end{equation}
where  $x_{ij} = x^T \cdot W_{ij} + b_{ij}$, $W \in R^{d \times m \times k}$ and $b_{ij} \in R^{m \times k}$ are the learned parameters.

The tanh activation function is defined as 
\begin{equation}\label{tanh}
f(x) = tanh(x) = \frac{2}{1 + \exp^{-2 \cdot x}} - 1
\end{equation}

\begin{figure}[h]
  \begin{minipage}[h]{0.49\linewidth}
    \center{\includegraphics[width=1\linewidth]{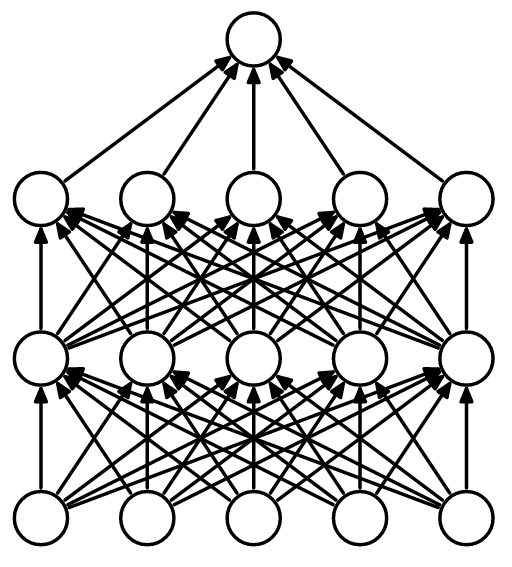}}
    \text{(a) Standard Neural Net}
  \end{minipage}
  \hfill
  \begin{minipage}[h]{0.49\linewidth}
    \center{\includegraphics[width=1\linewidth]{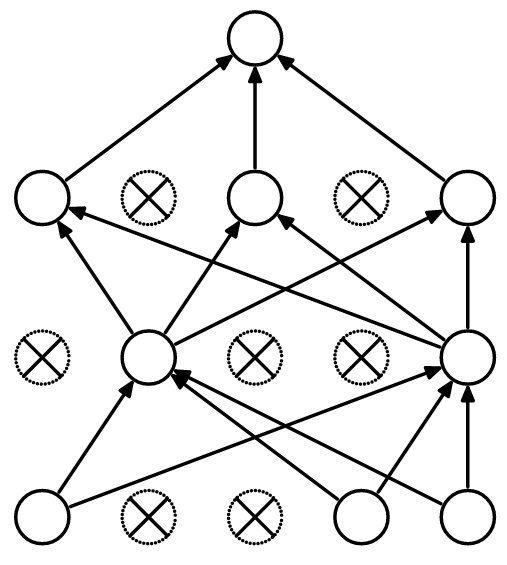}}
    \text{(b) Neural Net with Dropout}
  \end{minipage}
  \caption{An Example of Standard Neural Net on the Left and Neural Net with Dropout on the Right with 2 hidden layers~\cite{GHinton}}
  \label{fig2}
\end{figure}

Functions with dropout are used for reducinng overfitting by preventing complex co-adaptations on training data. This technique is known as regularization. Figure~\ref{fig2} demonstrate the difference between standatd neural net and neural net affter applying dropout~\cite{GHinton}.

\subsection{Constant Parameters of the Training Model}
 
We have used the network model with such paraeters, namely:

\begin{itemize}
  \item Response variable column is \textit{C785}
  \item Hidden layer sise is \textit{[50,50]}
  \item Epochs are \textit{500}
  \item Seed for random numbers is \textit{2}
  \item Adaptive learning rate is \textit{false}
  \item Initial momentum at the beginning of taining is \textit{0.9}
  \item Final momentum after the ramp is \textit{0.99}
  \item Input layer dropout ratio for improving generalization is \textit{0.2}
  \item Stopping criterion for classification error fraction on training data is \textit{disable}
  \item Early stopping based on convergence of stopping metric is \textit{3}
  \item Relative tolerance for metric-based stopping criterion is \textit{0.01}
  \item Compute variable impotances for input features is \textit{true}
   \item Sparse data handling is \textit{true}
  \item Rorce reproducibility on small data is \textit{true}
 
\end{itemize}

\subsection{Variable Parameters of the Training Model}

\begin{itemize}
  \item Activation function: \textit{Tanh, TanhWithDropout, Maxout, MaxoutWithDropout, Rectifier, RectifierWithDropout}
  \item Metric to use for early stopping: \textit{logloss,  misclassification, MAE, MSE, RMSE} and \textit{RMSLE}  
  \item Loss function:  \textit{Cross Entropy}
\end{itemize}

Loss function is a function that used to measure the degree of fit.
The cross entropy loss function for the distributions $p$ and $q$ over a given set is defined as follows:

\begin{equation}\label{H}
H(p, q) = H(p) + D_{KL}(p||q)
\end{equation}

where $H(p) $ is the entropy of $p$, and $D_{KL}(p||q)$ is the Kullback--Leibler divergence of $q$ from $p$ (also known as the relative entropy of $p$ with respect to $q$~\cite{CE}. Cross entropy is always larger than entropy.

\section{Experimental Results} \label{Experimental Results}

\begin{figure}[h!]
\centering
\includegraphics[width=3.5in]{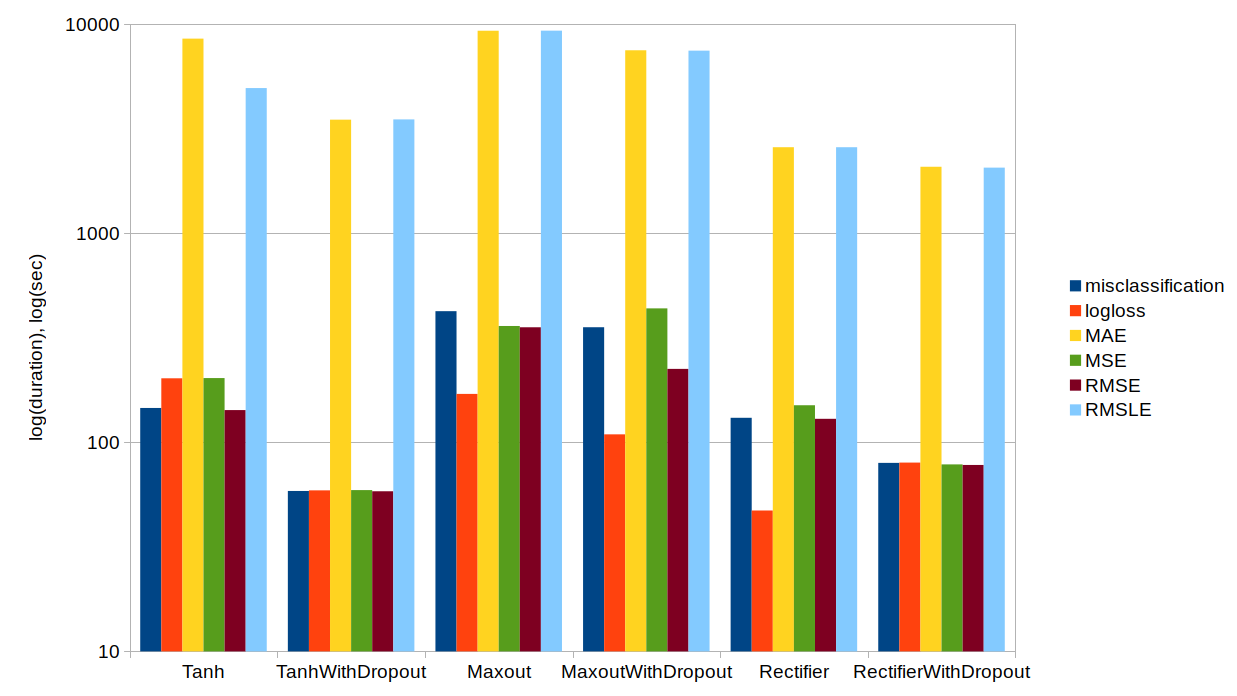}
\caption{Runtime of Learning Nets Different Architectures}
\label{fig3}
\end{figure}

We trained neural networks for classification problems on publicly available MNIST dataset of handwritten digits with use case in single-threaded mode. We found that  generalization  performance has very strong dependence on activation function and very slight dependence on stopping metric. Tables~\ref{duration} and~\ref{duration2} give give the values of runtime for the models used. Figure~\ref{fig3} shows the runtime  values on the logarithm scale obtained for these different architectures as training  progresses.

\begin{table}[h!] 
\renewcommand{\arraystretch}{1.5}
\caption{Runtime of Building Model, sec}
\label{duration}
\centering
\begin{tabular}{lllll}
\cline{1-4}
\multicolumn{1}{|c|}{\multirow{2}{*}{\textbf{\begin{tabular}[c]{@{}c@{}}Activation function\end{tabular}}}} & \multicolumn{3}{c|}{\textbf{Stopping metric}}   &  \\ \cline{2-4}

\multicolumn{1}{|c|}{} & \multicolumn{1}{c|}{\textit{misclassification}} & \multicolumn{1}{c|}{\textit{logloss}} & \multicolumn{1}{c|}{\textit{MAE}} & \\ \cline{1-4}

\multicolumn{1}{|l|}{Tanh}   & \multicolumn{1}{c|}{146.101}  & \multicolumn{1}{c|}{202.609}                          & \multicolumn{1}{c|}{8549.855} &  \\ \cline{1-4}

\multicolumn{1}{|l|}{TanhWithDropout} & \multicolumn{1}{c|}{58.559} & \multicolumn{1}{c|}{58.934}                          & \multicolumn{1}{c|}{3500.606} & \\ \cline{1-4}

\multicolumn{1}{|l|}{Maxout} & \multicolumn{1}{c|}{424.618} & \multicolumn{1}{c|}{170.543}                          & \multicolumn{1}{c|}{9326.579} & \\ \cline{1-4}

\multicolumn{1}{|l|}{MaxoutWithDropout} & \multicolumn{1}{c|}{355.756} & \multicolumn{1}{c|}{109.234}                          & \multicolumn{1}{c|}{7519.321} & \\ \cline{1-4}

\multicolumn{1}{|l|}{Rectifier} & \multicolumn{1}{c|}{131.102} & \multicolumn{1}{c|}{47.206}                          & \multicolumn{1}{c|}{2584.434} & \\ \cline{1-4}

\multicolumn{1}{|l|}{RectifierWithDropout} & \multicolumn{1}{c|}{79.791} & \multicolumn{1}{c|}{80.034}                          & \multicolumn{1}{c|}{2083.78} & \\ \cline{1-4}                                                                                         
\end{tabular}
\end{table}

\begin{table}[h!]
\renewcommand{\arraystretch}{1.5}
\caption{Continuation of Table \ref{duration} }
\label{duration2}
\centering
\begin{tabular}{lllll}
\cline{1-4}
\multicolumn{1}{|c|}{\multirow{2}{*}{\textbf{\begin{tabular}[c]{@{}c@{}}Activation function\end{tabular}}}} & \multicolumn{3}{c|}{\textbf{Stopping metric}}   &  \\ \cline{2-4}

\multicolumn{1}{|c|}{} & \multicolumn{1}{c|}{\textit{MSE}} & \multicolumn{1}{c|}{\textit{RMSE}} & \multicolumn{1}{c|}{\textit{RMSLE}} & \\ \cline{1-4}

\multicolumn{1}{|l|}{Tanh}   & \multicolumn{1}{c|}{203.094}  & \multicolumn{1}{c|}{142.718}                          & \multicolumn{1}{c|}{4959.809} &  \\ \cline{1-4}

\multicolumn{1}{|l|}{TanhWithDropout} & \multicolumn{1}{c|}{59.104} & \multicolumn{1}{c|}{58.351}                          & \multicolumn{1}{c|}{3511.679} & \\ \cline{1-4}

\multicolumn{1}{|l|}{Maxout} & \multicolumn{1}{c|}{360.663} & \multicolumn{1}{c|}{355.492}                          & \multicolumn{1}{c|}{9334.789} & \\ \cline{1-4}

\multicolumn{1}{|l|}{MaxoutWithDropout} & \multicolumn{1}{c|}{437.844} & \multicolumn{1}{c|}{224.742}                          & \multicolumn{1}{c|}{7490.342} & \\ \cline{1-4}

\multicolumn{1}{|l|}{Rectifier} & \multicolumn{1}{c|}{150.559} & \multicolumn{1}{c|}{129.651}                          & \multicolumn{1}{c|}{2584.347} & \\ \cline{1-4}

\multicolumn{1}{|l|}{RectifierWithDropout} & \multicolumn{1}{c|}{78.469} & \multicolumn{1}{c|}{77.964}                          & \multicolumn{1}{c|}{2064.164} & \\ \cline{1-4}                                                                                         
\end{tabular}
\end{table}

\begin{table}[h!]
\renewcommand{\arraystretch}{1.5}
\caption{Values of Training logloss}
\label{logloss}
\centering
\begin{tabular}{lllll}
\cline{1-4}
\multicolumn{1}{|c|}{\multirow{2}{*}{\textbf{\begin{tabular}[c]{@{}c@{}}Activation function\end{tabular}}}} & \multicolumn{3}{c|}{\textbf{Stopping metric}}   &  \\ \cline{2-4}

\multicolumn{1}{|c|}{} & \multicolumn{1}{c|}{\textit{misclassification}} & \multicolumn{1}{c|}{\textit{logloss}} & \multicolumn{1}{c|}{\textit{MAE}} & \\ \cline{1-4}

\multicolumn{1}{|l|}{Tanh}   & \multicolumn{1}{c|}{0.0759}  & \multicolumn{1}{c|}{0.0575}                          & \multicolumn{1}{c|}{0.0104} &  \\ \cline{1-4}

\multicolumn{1}{|l|}{TanhWithDropout} & \multicolumn{1}{c|}{1.2223} & \multicolumn{1}{c|}{1.1944}                          & \multicolumn{1}{c|}{1.3973} & \\ \cline{1-4}

\multicolumn{1}{|l|}{Maxout} & \multicolumn{1}{c|}{0.0537} & \multicolumn{1}{c|}{0.1734}                          & \multicolumn{1}{c|}{0.0009} & \\ \cline{1-4}

\multicolumn{1}{|l|}{MaxoutWithDropout} & \multicolumn{1}{c|}{0.397} & \multicolumn{1}{c|}{0.5358}                          & \multicolumn{1}{c|}{0.0986} & \\ \cline{1-4}

\multicolumn{1}{|l|}{Rectifier} & \multicolumn{1}{c|}{0.0577} & \multicolumn{1}{c|}{0.1298}                          & \multicolumn{1}{c|}{0.0216} & \\ \cline{1-4}

\multicolumn{1}{|l|}{RectifierWithDropout} & \multicolumn{1}{c|}{0.1449} & \multicolumn{1}{c|}{0.1449}                          & \multicolumn{1}{c|}{0.0865} & \\ \cline{1-4}                                                                                         
\end{tabular}
\end{table}

\begin{table}[h!]
\renewcommand{\arraystretch}{1.5}
\caption{Continuation of Table \ref{logloss}}
\label{logloss2}
\centering
\begin{tabular}{lllll}
\cline{1-4}
\multicolumn{1}{|c|}{\multirow{2}{*}{\textbf{\begin{tabular}[c]{@{}c@{}}Activation function\end{tabular}}}} & \multicolumn{3}{c|}{\textbf{Stopping metric}}   &  \\ \cline{2-4}

\multicolumn{1}{|c|}{} & \multicolumn{1}{c|}{\textit{MSE}} & \multicolumn{1}{c|}{\textit{RMSE}} & \multicolumn{1}{c|}{\textit{RMSLE}} & \\ \cline{1-4}

\multicolumn{1}{|l|}{Tanh}   & \multicolumn{1}{c|}{0.0575}  & \multicolumn{1}{c|}{0.0804}                          & \multicolumn{1}{c|}{0.0104} &  \\ \cline{1-4}

\multicolumn{1}{|l|}{TanhWithDropout} & \multicolumn{1}{c|}{1.1944} & \multicolumn{1}{c|}{1.1944}                          & \multicolumn{1}{c|}{1.3973} & \\ \cline{1-4}

\multicolumn{1}{|l|}{Maxout} & \multicolumn{1}{c|}{0.0537} & \multicolumn{1}{c|}{0.0537}                          & \multicolumn{1}{c|}{0.0009} & \\ \cline{1-4}

\multicolumn{1}{|l|}{MaxoutWithDropout} & \multicolumn{1}{c|}{0.2950} & \multicolumn{1}{c|}{0.4653}                          & \multicolumn{1}{c|}{0.0986} & \\ \cline{1-4}

\multicolumn{1}{|l|}{Rectifier} & \multicolumn{1}{c|}{0.0577} & \multicolumn{1}{c|}{0.0596}                          & \multicolumn{1}{c|}{0.0216} & \\ \cline{1-4}

\multicolumn{1}{|l|}{RectifierWithDropout} & \multicolumn{1}{c|}{0.1449} & \multicolumn{1}{c|}{0.1449}                          & \multicolumn{1}{c|}{0.0865} & \\ \cline{1-4}                                                                                         
\end{tabular}
\end{table}

\begin{figure}[h!]
\centering
\includegraphics[width=3.5in]{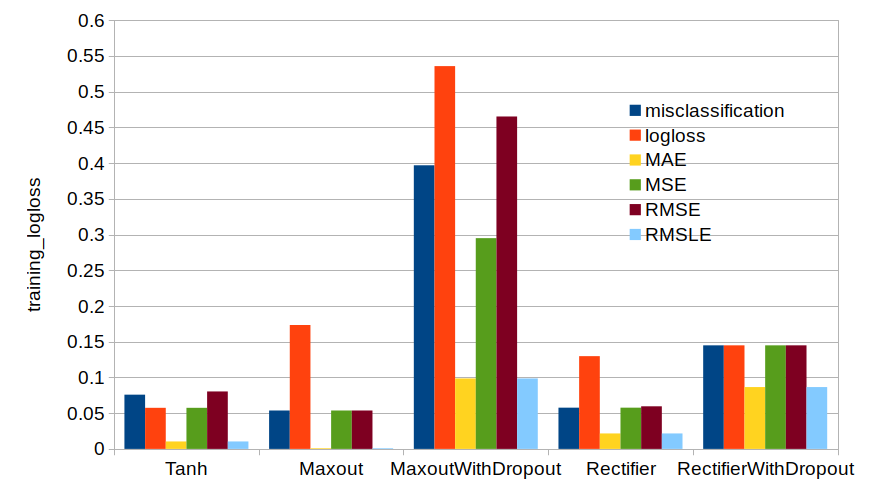}
\caption{Training Logloss of Learning Nets Different Architectures}
\label{fig4}
\end{figure}

Tables~\ref{logloss} and~\ref{logloss2} contain the information of training errors  for various activation function and stopping metrics. Figure~\ref{fig4} demonstrates the effectiveness of using tanh activation function for all stopping metrics that considered in this paper. In the case of the learning net based on the tanh activation function, MAE and RMSLE stopping metric hase achieved the logloss value of 0.0104. This architectures demonstrate better training prediction ability than others but take much time for building model.

\begin{figure}[h!]
\centering
\includegraphics[width=3.5in]{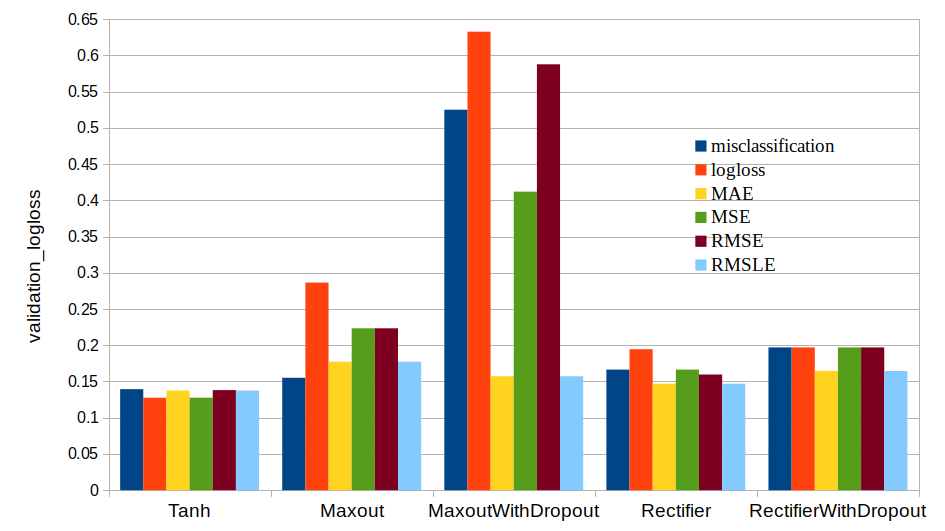}
\caption{Validation Logloss of Learning Nets Different Architectures}
\label{fig5}
\end{figure}

In order to find the best neural net architecture for digits recognition just needs to look at the behavior of models on unknown data  should be checked. Tables~\ref{Vlogloss} and~\ref{Vlogloss2} give the information that can help us to find out the net architecture that provides the best performance in case of using single-threaded mode. Figure~\ref{fig5} shows the validation error rates  for different architectures that are considered here. We see, the best digits recognition results were achieved in the case of tanh activation function. The type of stopping metric is very slightly effects on the values of the validation error but it does very much on the runtime of building model.

\begin{table}[h!]
\renewcommand{\arraystretch}{1.5}
\caption{Values of Validation logloss}
\label{Vlogloss}
\centering
\begin{tabular}{lllll}
\cline{1-4}
\multicolumn{1}{|c|}{\multirow{2}{*}{\textbf{\begin{tabular}[c]{@{}c@{}}Activation function\end{tabular}}}} & \multicolumn{3}{c|}{\textbf{Stopping metric}}   &  \\ \cline{2-4}

\multicolumn{1}{|c|}{} & \multicolumn{1}{c|}{\textit{misclassification}} & \multicolumn{1}{c|}{\textit{logloss}} & \multicolumn{1}{c|}{\textit{MAE}} & \\ \cline{1-4}

\multicolumn{1}{|l|}{Tanh}   & \multicolumn{1}{c|}{0.1394}  & \multicolumn{1}{c|}{0.1276}                          & \multicolumn{1}{c|}{0.1375} &  \\ \cline{1-4}

\multicolumn{1}{|l|}{TanhWithDropout} & \multicolumn{1}{c|}{1.1879} & \multicolumn{1}{c|}{1.1659}                          & \multicolumn{1}{c|}{1.3838} & \\ \cline{1-4}

\multicolumn{1}{|l|}{Maxout} & \multicolumn{1}{c|}{0.1551} & \multicolumn{1}{c|}{0.2865}                          & \multicolumn{1}{c|}{0.1773} & \\ \cline{1-4}

\multicolumn{1}{|l|}{MaxoutWithDropout} & \multicolumn{1}{c|}{0.5250} & \multicolumn{1}{c|}{0.6327}                          & \multicolumn{1}{c|}{0.1572} & \\ \cline{1-4}

\multicolumn{1}{|l|}{Rectifier} & \multicolumn{1}{c|}{0.1664} & \multicolumn{1}{c|}{0.1945}                          & \multicolumn{1}{c|}{0.147} & \\ \cline{1-4}

\multicolumn{1}{|l|}{RectifierWithDropout} & \multicolumn{1}{c|}{0.1970} & \multicolumn{1}{c|}{0.1970}                          & \multicolumn{1}{c|}{0.1645} & \\ \cline{1-4}                                                                                         
\end{tabular}
\end{table}

\begin{table}[h!]
\renewcommand{\arraystretch}{1.5}
\caption{Continuation of Table \ref{Vlogloss}}
\label{Vlogloss2}
\centering
\begin{tabular}{lllll}
\cline{1-4}
\multicolumn{1}{|c|}{\multirow{2}{*}{\textbf{\begin{tabular}[c]{@{}c@{}}Activation function\end{tabular}}}} & \multicolumn{3}{c|}{\textbf{Stopping metric}}   &  \\ \cline{2-4}

\multicolumn{1}{|c|}{} & \multicolumn{1}{c|}{\textit{MSE}} & \multicolumn{1}{c|}{\textit{RMSE}} & \multicolumn{1}{c|}{\textit{RMSLE}} & \\ \cline{1-4}

\multicolumn{1}{|l|}{Tanh}   & \multicolumn{1}{c|}{0.1276}  & \multicolumn{1}{c|}{0.1381}                          & \multicolumn{1}{c|}{0.1375} &  \\ \cline{1-4}

\multicolumn{1}{|l|}{TanhWithDropout} & \multicolumn{1}{c|}{1.1659} & \multicolumn{1}{c|}{1.1659}                          & \multicolumn{1}{c|}{1.3838} & \\ \cline{1-4}

\multicolumn{1}{|l|}{Maxout} & \multicolumn{1}{c|}{0.2234} & \multicolumn{1}{c|}{0.2234}                          & \multicolumn{1}{c|}{0.1773} & \\ \cline{1-4}

\multicolumn{1}{|l|}{MaxoutWithDropout} & \multicolumn{1}{c|}{0.4120} & \multicolumn{1}{c|}{0.5877}                          & \multicolumn{1}{c|}{0.1572} & \\ \cline{1-4}

\multicolumn{1}{|l|}{Rectifier} & \multicolumn{1}{c|}{0.1664} & \multicolumn{1}{c|}{0.1596}                          & \multicolumn{1}{c|}{0.1470} & \\ \cline{1-4}

\multicolumn{1}{|l|}{RectifierWithDropout} & \multicolumn{1}{c|}{0.1970} & \multicolumn{1}{c|}{0.1970}                          & \multicolumn{1}{c|}{0.1645} & \\ \cline{1-4}                                                                                         
\end{tabular}
\end{table}

\section{Conclusion} \label{Conclusion}
H2O is widely used open source deep learning platform.In this paper, we present the results of testing neural networks architectures on H2O platform for various activation functions, stopping metrics, and other parameters of machine learning algorithm. It was demonstrated for the use case of MNIST database of handwritten digits in single-threaded mode that blind selection of these parameters can hugely increase (by 2-3 orders) the runtime without the significant increase of precision. This result can have crucial influence for opitmization of available and new machine learning methods, especially for image recognition problems. 

During the process of testing H2O, we found out that  generalization  performance has very strong dependence on activation function and very slight dependence on stopping metric. The best results of recognition digits were achieved in case of using nets architecture based on tanh activation function, logloss and MSE  stopping metrics.

\end{document}